%% file: main.tex
\documentclass[letterpaper, 10 pt, conference]{ieeeconf}  

\IEEEoverridecommandlockouts                              

\usepackage{graphics} 
\usepackage{epsfig} 
\usepackage{amsmath} 

\usepackage{multirow}

\usepackage{amsfonts}    
\usepackage{booktabs}
\usepackage{xcolor}

\usepackage{algorithm}   
\usepackage[noend]{algpseudocode} 

\floatname{algorithm}{Procedure}

\usepackage{amssymb}
\usepackage{pifont}
\newcommand{\cmark}{\ding{51}}%
\newcommand{\xmark}{\ding{55}}

\makeatletter
\let\NAT@parse\undefined
\makeatother
\usepackage{hyperref}
\hypersetup{colorlinks,citecolor={blue}} 

\usepackage{listings}

\title{Human Following in Mobile Platforms with Person Re-Identification}

\author{%
  Mario Srouji$^\dagger$,
  Yao-Hung Hubert Tsai$^\dagger$, Hugues Thomas, Jian Zhang\thanks{ $^\dagger$equal contribution. Apple, contact:
  \texttt{\{msrouji,yaohung\_tsai,hthomas23,jianz\}@apple.com}}
}

\begin{document}

\maketitle
\thispagestyle{empty}
\pagestyle{empty}

\begin{abstract}
    Human following is a crucial feature of human-robot interaction, yet it poses numerous challenges to mobile agents in real-world scenarios. Some major hurdles are that the target person may be in a crowd, obstructed by others, or facing away from the agent. To tackle these challenges, we present a novel person re-identification module composed of three parts: a 360-degree visual registration, a neural-based person re-identification using human faces and torsos, and a motion tracker that records and predicts the target person's future position. Our human-following system also addresses other challenges, including identifying fast-moving targets with low latency, searching for targets that move out of the camera's sight, collision avoidance, and adaptively choosing different following mechanisms based on the distance between the target person and the mobile agent. Extensive experiments show that our proposed person re-identification module significantly enhances the human-following feature compared to other baseline variants.
\end{abstract}

\input{sections/intro}
\input{sections/related}
\input{sections/method}

\input{sections/experiments}
\input{sections/conclusion}

{
\small
\bibliography{ref_corl}
\bibliographystyle{plain}
}

\end{document}

%% file: sections/intro.tex
\section{Introduction}
\label{sec:intro}

Recent advancements in artificial intelligence have made it possible for humans to cooperate with mobile agents. This cooperation includes autonomous delivery platforms, service-based autonomous agents taking care of elders in hospitals, household agents cleaning floors, and telepresence agents creating new communication methods. These robots cooperate with humans through various types of interactions, and we argue that the human following feature is one of the most crucial.

The objective of this paper is to design a human following system that meets the following requirements: 1) the agent can follow a targeted person, even if they are moving quickly or away from the agent, 2) the agent can search for the person if they move out of the camera's sight, 3) the agent can avoid obstacles while following the target, and 4) the agent can still identify the target even when they are in a crowd. Our human following approach is able to address all of the above requirements, and the core is a novel person re-identification module.

Our person re-identification module consists of three main components. First, we introduce a 360-degree registration process to capture different angles of the target, which is an improvement over standard registration processes in prior literature~\cite{chen2017person,chen2017integrating,algabri2020deep} that only take a targeted person's front and back appearance into account. This process greatly helps us re-identify the target even when the target is side-facing the camera. Second, our method adopts different identification models for different body parts (face and torso) to re-identify the targeted person, unlike prior literature~\cite{koide2016identification,gross2017roreas,gupta2016novel} that use the entire body's appearance. This approach combines the best of both worlds: primarily using faces for re-identification, since faces carry the most distinctive feature among body parts~\cite{wu2019deep}; secondarily using torso for re-identification when faces are occluded or not present in the camera's sight. Third, we consider a motion tracker (e.g., Kalman filter)~\cite{Bewley2016_sort} to predict the target's future position, which helps to improve tracking when the target is occluded or far away. In addition, we show that the motion tracker can help us reduce the latency of running the person re-identification module, allowing our mobile agent to still track the target when they are moving quickly.

Our human-following system also consists of several other modules, including collision avoidance with local planning and path planning, a searching algorithm that locates the targeted person when out of the camera's sight, and a following mechanism based on a RGBD camera and a fish-eye camera. By integrating all of these modules, our system controls the mobile agent to follow the targeted person and keep them in the camera's line of sight, while also preventing collisions with obstacles. It is important to note that our system runs directly on the agent's mobile device for on-device mobile computation. This is different from prior works~\cite{scholtz2003theory,steinfeld2006common,goetz2003matching,waldherr2000gesture,hirai2003visual,chen2017person,gupta2016novel,chen2017integrating,algabri2020deep} that run human following systems on a server and send control commands from the server to the mobile agent.

As a summary, in this paper, we present a novel human-following system that incorporates a person re-identification module. We conducted extensive experiments and user studies to demonstrate that our system outperforms other baseline approaches in two different conditions: when a targeted person is walking to random markers in an environment and when a targeted person is following a course. The environments contain obstacles, and there are always two people in the environment: one is the target person, and the other is an interferer. Our system demonstrates strong performance in various metrics, including the average speed of the agent following the target person, the average following distance between the agent and the target person, the average distance to obstacles, the number of times the agent loses the target person, and the number of times the agent follows the wrong person.

%% file: sections/related.tex
\section{Related Work}
\label{sec:related}

Earlier human following techniques~\cite{scholtz2003theory,steinfeld2006common,goetz2003matching,waldherr2000gesture,yao2017monocular,hirai2003visual} can be understood as performing detection and tracking for humans. Recent approaches~\cite{chen2017person,gupta2016novel,chen2017integrating,algabri2020deep} consider person re-identification as an additional module on top of detection and tracking. In the following, we first discuss the approaches without person re-identification, and then we discuss the approaches with person re-identification.

\input{fig_tex/person_reid}

As some of the earliest attempts, Nagumo and Ohya~\cite{nagumo2001human} asked the targeted person to carry LED lights for the mobile agent to detect and track. Schlegal {\it et al.}~\cite{schlegel1998vision} proposed to use the human's contour and color histogram as the human following signal. Hirai and Mizoguchi~\cite{hirai2003visual} used the human back and shoulder as the human following signal, and Hu {\it et al.}~\cite{hu2013design} used leg appearance as the human following signal. Instead of using only cameras on the mobile agent, Marioka {\it et al.}~\cite{morioka2004human} considered external cameras to improve detection and tracking. Nonetheless, using external cameras lowers the practical applicability for the human following feature. Other than visual features, Han and Jin~\cite{han2015sound} explored the usage of the audio signal for human following. Since these approaches do not consider person re-identification, their human following feature can easily fail when the targeted person 1) is occluded or 2) intersects with another individual.

For the approaches with person re-identification ability, Gupta {\it et al.}~\cite{gupta2016novel} and Gross {\it et al.}~\cite{gross2017roreas} presented the use of template matching to identify the targeted person. Koide {\it et al.}~\cite{koide2016identification} used the edges, color, and texture of the targeted person's clothes as the identification features. Nonetheless, these approaches use pre-deep-learning vision features such as SURF~\cite{bay2006surf}, which have been shown to be less effective nowadays~\cite{lecun2015deep}. To address this concern, work~\cite{chen2017person,chen2017integrating,algabri2020deep} presented the use of deep-learning features as identification features. Our approach is most similar to these works, yet we consider 1) a different registration process (360-degree registration), 2) person identification models by multiple parts (faces and torsos), and 3) the integration with a motion tracker.

%% file: fig_tex/person_reid.tex
\begin{figure*}[t!]
    \centering
    \includegraphics[width=0.95\linewidth]{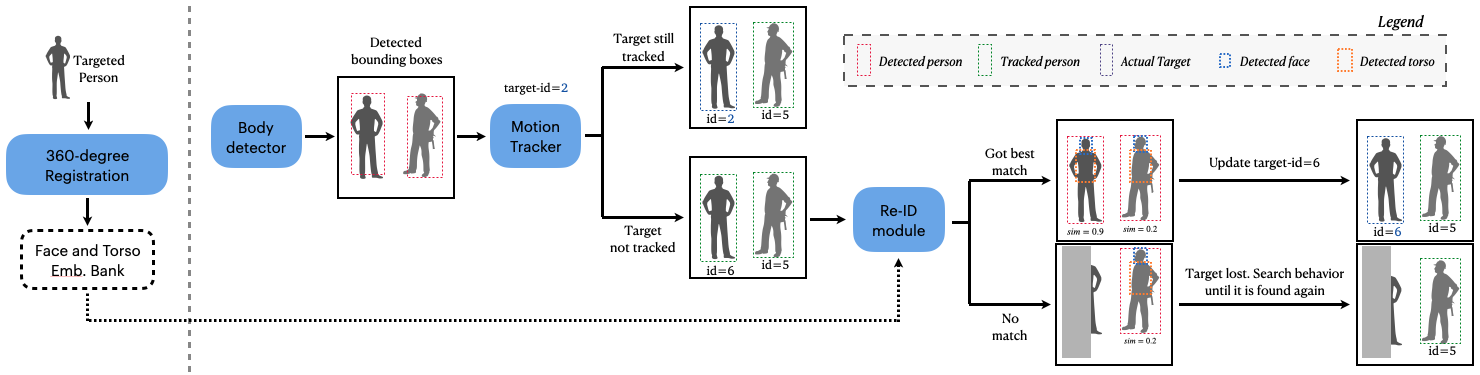}
    \caption{
    We register a target person using our 360-degree registration module. Our tracking and re-identification pipeline is then able to find the targeted person and track them among other individuals. Whenever the target is lost, our re-id module comes into action. It detects the torso and face (if available) of each detected person and computes corresponding embeddings. The target is identified by comparing these embeddings to the saved feature bank. If the target cannot be found, the robot enters search mode.
    }
    \label{fig:person_reid}
\end{figure*}

%% file: sections/method.tex
\section{Proposed System}
\label{sec:proposed}

In this section, we will first describe the person re-identification module, and then we will discuss our human following system with the remaining modules and algorithms.

\subsection{Person Tracking and Re-Identification Pipeline}

We propose a person tracking and re-identification pipeline that includes three components: a 360-degree registration process, a body detection and tracking pipeline, and a person identification model that uses multiple body parts (face and torso). We depict this pipeline in Figure~\ref{fig:person_reid}.

{\bf Body detection model.} For every frame, our system first runs a body and pose detector that provides the body bounding boxes and the poses of all humans in the image. We leverage the open source software provided by \cite{zhou2019objects}\footnote{\textit{\urlstyle{sf}\url{https://github.com/xingyizhou/CenterNet}}}, which provides good detection performances.

{\bf Motion Tracker.} We use a standard open-source Kalman filter motion tracker ~\cite{Bewley2016_sort}\cite{welch1995introduction}\footnote{\textit{\urlstyle{sf}\url{https://github.com/abewley/sort}}} which tracks the position of the detected body bounding boxes. Each tracked bounding box is assigned a consistent ID across frames allowing tracking of the target even with noisy detections or when the target is occasionally not detected for a few frames.

{\bf Face and Torso Identification Models.} To distinguish the targeted individual from others, we utilize three additional neural-based machine learning models: a face detector that provides face bounding boxes in the current frame; a face embedding model that provides a face embedding on a cropped face bounding box; and a torso embedding model that provides a torso embedding on a cropped torso bounding box. 

\input{fig_tex/face_torso_id_model}

We demonstrate the functionality of the face and torso identification models in Figure~\ref{fig:face_torso_id_models}. First, given the initially detected body poses, we estimate the person's face and torso bounding boxes. We draw a square bounding box on the face with its width and height being the distance between the left and right ears. Then, we draw a rectangular bounding box on the torso with its width and height covering the left and right shoulders and hips. However, we have observed that the pose-induced face bounding boxes may not always be accurate, especially when the person is not facing the camera. In such cases, the pose-induced face bounding boxes may be falsely detected. On the other hand, the pose-induced torso bounding boxes are often accurate.

Then, we obtain better face bounding boxes for all individuals using the unofficial open-source implementation \footnote{\textit{\urlstyle{sf}\url{https://github.com/chenjun2hao/CenterFace.pytorch}}} of CenterFace \cite{CenterFace}. Note that if an individual is not facing the camera, their face may not be detected, resulting in a lower number of face bounding boxes than body bounding boxes. These detected face bounding boxes are matched with the pose-induced face bounding boxes based on their intersection of union (IoU) values. If the corresponding IoU value is greater than $0.75$, we assign the detected face bounding boxes (from the face detection model) to a person. 

Eventually, we produce the face and torso embeddings by passing the face and torso bounding boxes to two different models: FaceNet \cite{schroff2015facenet} and Object-reID \cite{zheng2019joint}. Again, we use open-source repositories
\footnote{\textit{\urlstyle{sf}\url{https://github.com/timesler/facenet-pytorch}}}
\footnote{\textit{\urlstyle{sf}\url{https://github.com/layumi/Person_reID_baseline_pytorch}}} 
for reproducibility. Note that a person may not have a face embedding if their face is not present in the image. In summary, our models for identifying faces and torsos will provide the bounding box for each person, as well as the torso embedding. Additionally, the face embedding may be provided as an optional feature.

{\bf 360 Registration Process.} Initially, we choose a target to follow, and collect its face and torso embeddings using a 360-registration process as shown in Figure~\ref{fig:registration}. The targeted person is asked to turn around in front of our mobile agent. The process usually takes about 20 seconds to complete, during which our models process hundreds of images. We then randomly select 100 face embeddings and 100 torso embeddings to form the feature bank.
 
\input{fig_tex/registration}

{\bf Re-identification module.} At the initial steps, the tracker is not yet aware of which ID corresponds to the target individual. The tracker might also lose the target due to occlusions or the person going out of sight. In those cases, the re-identification (re-id) module is called. It compares the target's embeddings stored in the feature bank with the face and torso embeddings of all the detected individuals in the scene.

To perform the comparison, we initially compute the average of the embeddings within the bank. Then, we measure the cosine similarity between the individual's face and torso embeddings and the calculated average bank embeddings. A similarity score is computed as the maximum value between the face and the torso cosine similarities. If the highest score is above a fixed threshold ($sim > 0.8$), we establish the corresponding person as the target. Otherwise, our following system starts its searching behavior, as described in the next section.

On our mobile platform, we can run the person re-id module at 8 fps. This low frequency makes it hard to follow fast-moving targets. This is why we rely on the motion tracker most of the time, and make calls to the re-id module only when necessary. By doing so, we can match the camera frame rate up to 30 fps, and significantly reduce the reaction time of our following pipeline for the majority of the time.

\subsection{Proposed Human Following System}
\label{subsec:human_following_system}

Our system for human following, illustrated in Figure~\ref{fig:action}, consists of our motion tracking and re-id modules, in addition to other modules including a dual camera setup, a local planner and collision avoidance module, and a search behavior to retrieve missing targets. The goal of the system is to follow a targeted person by controlling the mobile agent's movements (which include moving forward, moving back, rotating left, rotating right, etc).

{\bf Dual camera setup.} The challenges for a following system are different depending on the distance to its target. If the target is close to the system, it may go out of sight easily by walking to the side of the mobile agent. If it is far from the system, obstacles might be in the way, introducing occlusions or blocked paths. Therefore we use a short-range fish-eye RGB camera with a large field of view and a long-range RGBD camera with a more narrow field of view, but providing depth information. As suggested by prior research ~\cite{gupta2016novel,morioka2004human,algabri2020deep, hirai2003visual,chen2017person,chen2017integrating,yao2017monocular} depth information is very valuable to follow humans as it allows for path planning. The switch between the two cameras is triggered by the measured depth of the person when using the RGBD camera or by the size of the bounding box when using the fish-eye camera. If the bounding box height is smaller than $45\%$ of the image, we switch to the RGBD camera. If the person is closer than $1.5$ meters to the camera, we switch to the fish-eye camera.

{\bf Navigation.} When using the fish-eye camera, we use a very simple visual servoing method, which produces a control command based on the bounding box position in the image, aiming at centering this bounding box, and maximizing the height of the bounding box to a certain extent. When using the RGBD camera, we are able to estimate the depth of the target and thus its $(x, y)$ position relative to the system. Therefore the control command can be produced directly with the local planner described below, using this local goal.

{\bf Local Planner and Collision Avoidance.} To ensure the mobile agent's safety, we take into account any obstacles that may be present between the agent and its target. This is done through the use of the SAFER local planner and collision avoidance algorithm~\cite{srouji2022safe}. This algorithm can either take a control command or a local goal as input and it outputs a safe control command that avoids obstacles.

\input{fig_tex/action}

{\bf Search behavior.} Thanks to the fish-eye camera and visual servoing, it is rare to lose the person when it walks to the side of the mobile agent. Nevertheless, this still might happen, and occlusions might also lead to the system losing its target. In this case, the system enters into search mode. When the system is using the RGBD camera, the last known position of the target is kept as the local goal for a short period of time (e.g. 2 seconds). After that, or when using the fish-eye camera, our mobile agent will stop moving forward and rotate in the direction where the target was last seen. This allows our mobile agent to have a better chance at re-finding the target.

%% file: fig_tex/face_torso_id_model.tex
\begin{figure*}[t!]
    \centering
    \includegraphics[width=0.95\linewidth]{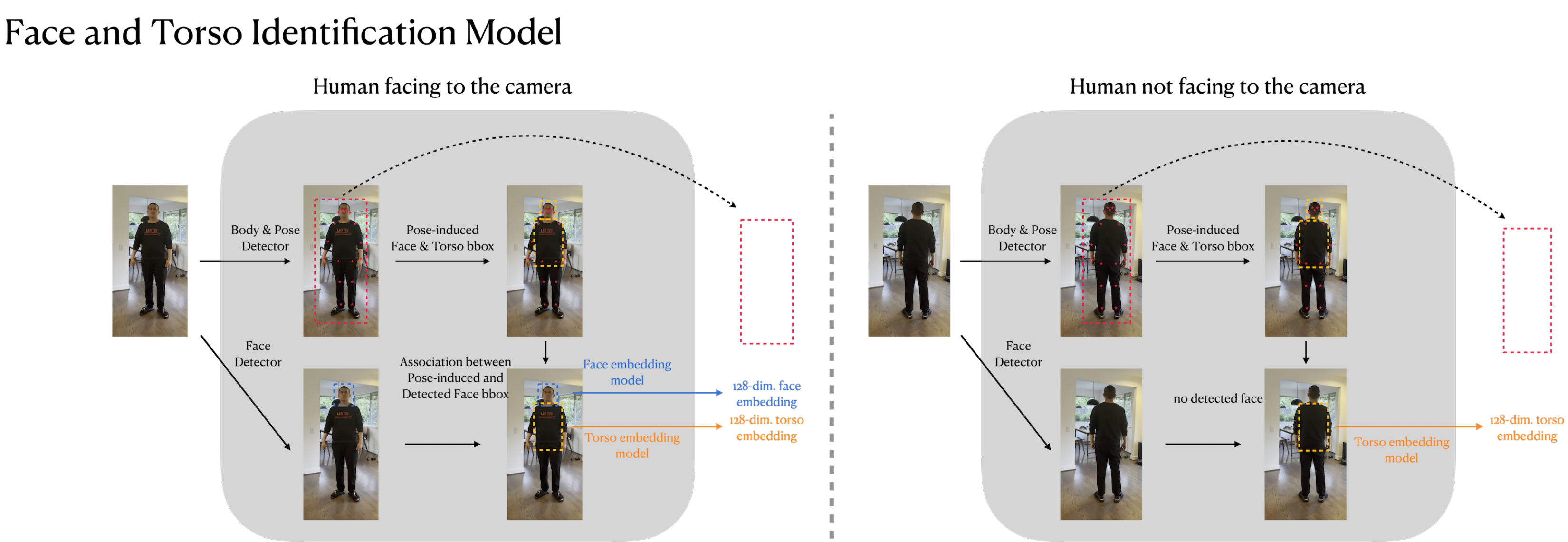}
    \caption{Left: when human is facing to the camera. Right: when human is not facing to the camera. When there is a face detected, our face and torso identification model will return 1) human bounding box, 2) face embedding, and 3) torso embedding. When there is no face detected, our face and torso identification model will only return 1) human bounding box and 2) torso embedding.}
    \label{fig:face_torso_id_models}
\end{figure*}

%% file: fig_tex/registration.tex
\begin{figure*}[t!]
    \centering
    \includegraphics[width=0.95\linewidth]{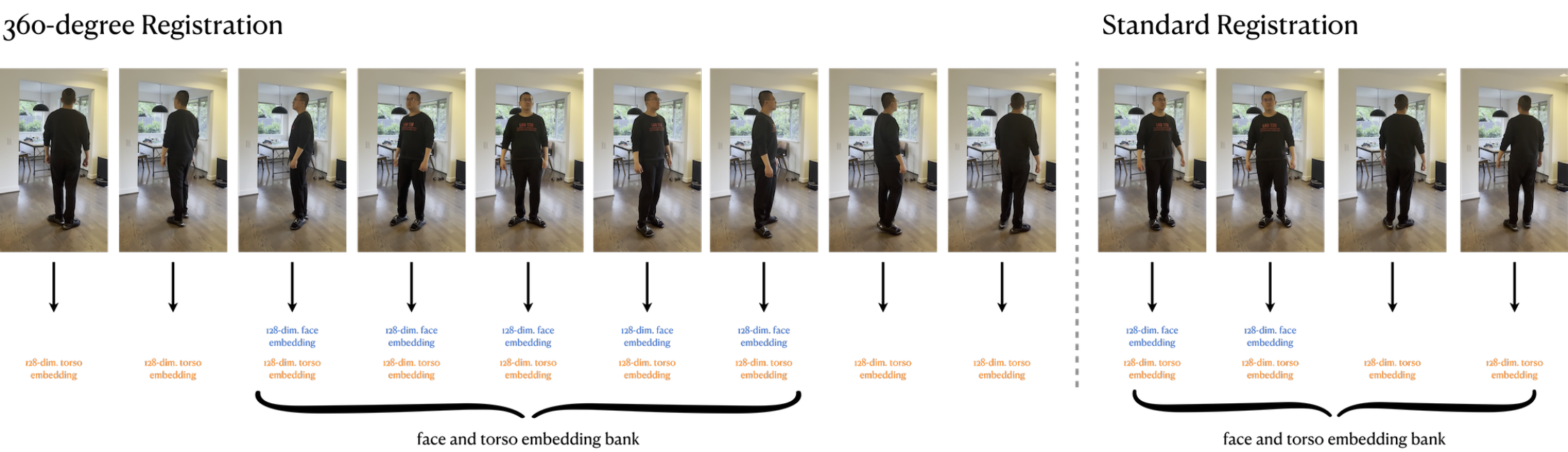}
    \caption{360-degree and standard registration process for registering a targeted person's face and torso embeddings. 360-degree registration process captures the embeddings from different angles of the target, while standard registration process captures only the front-facing and back-facing angels.}
    \label{fig:registration}
\end{figure*}

%% file: fig_tex/action.tex
\begin{figure}[b]
    \centering
    \includegraphics[width=0.95\linewidth]{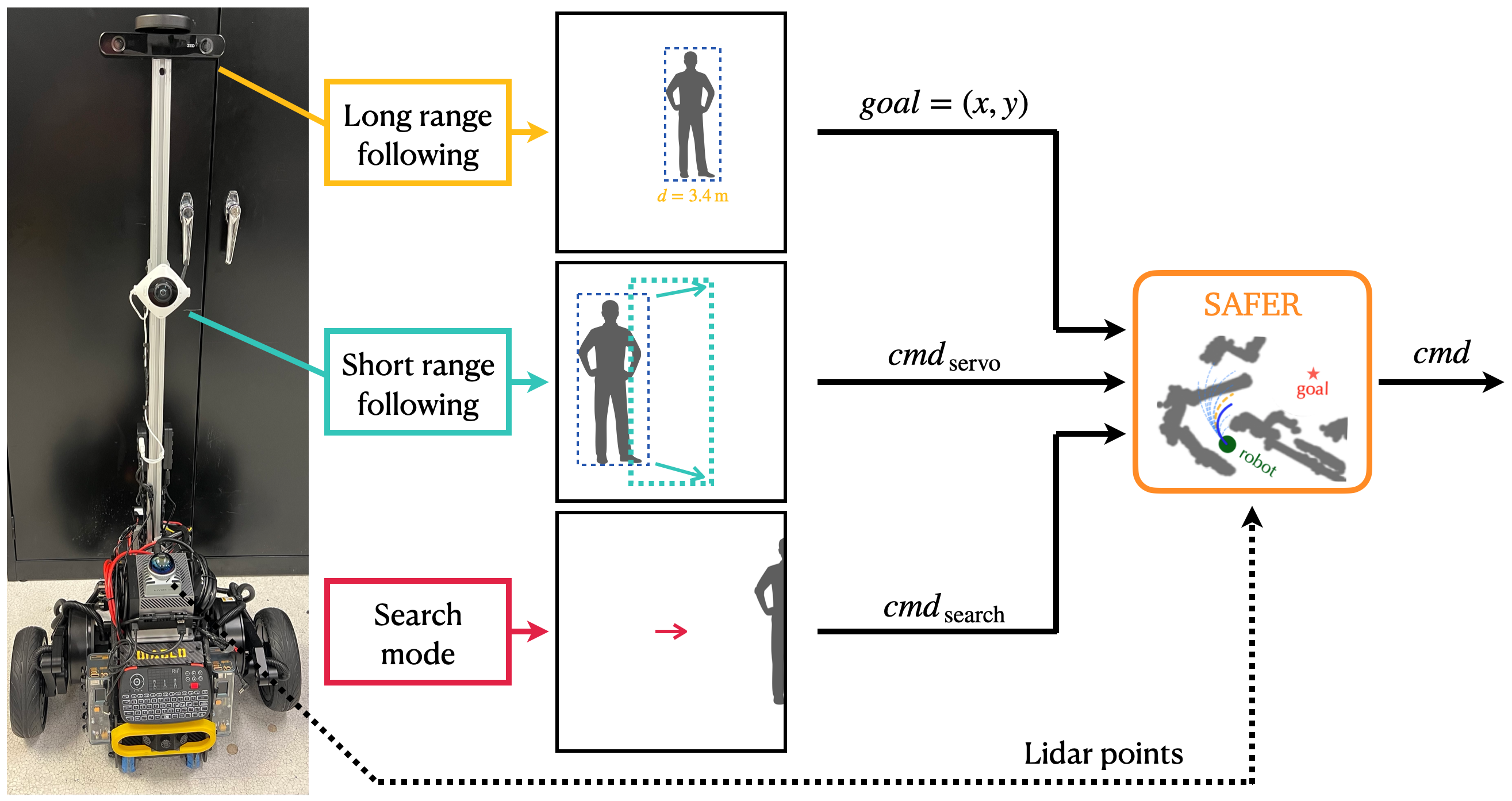}
    \caption{
    Our mobile agent uses a dual-camera setup. A fish-eye RGB camera for short-range following using simple visual servoing and a RGBD camera for longer-range following using a local navigation goal. We leverage SAFER, a local planer and collision avoidance module \cite{srouji2022safe} to handle safe navigation towards the target. In case the robot is in search mode, it will turn in the direction where the target was last seen.
    }
    \label{fig:action}
\end{figure}

%% file: sections/experiments.tex
\section{Experiments}
\label{sec:exp}

Our system, designed to track humans, is installed in a mobile agent that includes a base from Diablo Robotics~\cite{Diablo}, Nvidia Jetson Orin~\cite{nvidia} compute, a J5 Create JVCU360 camera~\cite{j5create}, a Zed camera~\cite{zed}, and a Livox Mid-360 lidar sensor~\cite{livox}. We conducted experiments in indoor office environments where the mobile agent followed a target person. We tested two situations: the target person walking to random markers and the target person following a course. The environments included obstacles, and always had two people present, one being the target person and the other being the interferer.

\subsection{Comparison Approaches} 

For our experiments, we began by considering human-following systems without person re-identification. We first looked at the variant that only tracks the human, which is similar to the prior method~\cite{yoshimi2006development}. Next, we looked at variants that use face-only or torso-only identification models. It's worth noting that the variant that only considers torso-only identification models is similar to the prior method~\cite{chen2017integrating}, which uses a full-body identification module (e.g., clothing color, height, gait) for human tracking.

We then considered our method without using a motion tracker, and finally, we looked at our method that uses only visual servo or only path planning to drive the mobile agent. Our goal in designing these baseline variants was to show the importance of 1) person re-identification in the human-following feature and 2) the individual parts of our person re-identification system (e.g., face-torso identification models, motion tracker, mobile agent driving by visual servo or path planning). See Table~\ref{tbl:baselines} for our summary.

\input{tbl_tex/baselines.tex}

\input{tbl_tex/results}

\subsection{Metrics}
\label{subsec:metric_questions}

To determine how to successfully follow a human, we conducted user studies to establish metrics. Our experiments involved 5 participants walking to 15 random markers (75 trials in total) and 5 participants following 3 different courses (15 trials in total). These trials were conducted in various office environments, each containing obstacles such as chairs, desks, cabinets, or walls, as well as an interferer who randomly walked in the same environment to interrupt the agent following the target person.

To measure success, we report 1) the average speed of the agent when following the target person, 2) the average distance between the agent and followed person, 3) the average distance to obstacles according to the Livox Mid-360 lidar~\cite{livox}, 4) the number of times the agent loses the target person, and 5) the number of times the agent follows the wrong person. Additionally, we created a questionnaire and asked participants to rate their experience on a scale from 1 to 10, reporting the average score for each question.

The questions are
\begin{enumerate}
    \item[Q1] [Safety], how safe did you feel while the robot was following you? (With 1 being not safe at all and 10 being very safe)
    \item[Q2] [Collision to Participants], how often did you feel the robot might collide with you? (With 1 being never and 10 being every time)
    \item[Q3] [Collision to other Objects], how often did you feel the robot might collide with another object? (With 1 being never and 10 being every time)
    \item[Q4] [Following Accuracy], how accurately did the robot follow you? (With 1 being not accurate at all and 10 being very accurate)
    \item[Q5] [Navigation Smoothness], how smoothly did the robot navigate to follow you? (With 1 being not smoothly at all and 10 being at very smoothly)
\end{enumerate}

\input{tbl_tex/quetionairs}

\subsection{Quantitative evaluation}

The quantitative results of our experiments are compiled in Table~\ref{tbl:results}. First, we compared methods with (Ours) and without re-identification (re-id) models (Ours\_w/o\_reid). We found that using re-id models can improve the speed of the agent, reduce the following distance to the person being followed, and significantly decrease the number of lost targets and incorrect following. This observation suggests that using re-id models can greatly improve the human following feature.

Similarly, when comparing methods with (Ours) and without the motion tracker (Ours\_w/o\_motion), we observe that the speed of the agent when following a person is greatly improved after using the motion tracker. We argue that this is because, without using the motion tracker, the agent must perform re-identification with re-id models most of the time, resulting in high latency.

Then, we discuss methods using different re-id models. Comparing Ours\_w/o\_torso and Ours\_w/o\_face, we find that using face re-id models leads to higher average speed of the agent, lower average following distance, and far less following of the wrong person. This can be seen when comparing Ours\_w/o\_torso, which results in a higher number of lost targets, but a lower number for following an incorrect person when the agent is walking to a random marker. Ours\_w/o\_face has less overall lost targets, but all of the lost targets resulted in an incorrect following. We argue that this is because face is a stronger feature for person identification, however since many times the person may be facing away from the agent, the torso may be the only identification feature for the agent. Combining the best of both worlds, Ours improves over Ours\_w/o\_torso and Ours\_w/o\_face, suggesting that both torso and face re-id models are crucial for human following.

Lastly, we discuss methods using different approaches (visual servo, path planning, or both) for the agent to take actions. Comparing Ours\_w/o\_visualservo and Ours\_w/o\_pathplanning, the main difference is that the average distance to obstacles is much larger when using path planning only, and the average speed is lower. In other words, path planning is a more conservative approach for collision avoidance than visual servo. On the other hand, Ours determines when to use visual servo or path planning depending on the distance of the target person to the agent. We find this to be a good combination of both approaches.

\subsection{User study}

In addition to raw measurements, we designed a user study to evaluate how user would feel around our following system. The questions listed in Section~\ref{subsec:metric_questions} aim at evaluating different aspects of the following experience.

Questions 1, 2, and 3 are overall focusing on the feeling of safety that participants got from the system. Overall, the scores reflect a pretty good feeling of safety for all of the ablated variants of our method except for Ours\_w/o\_pathplanning. This observation suggests that using path planning for the agent's action is crucial for avoiding collisions. The best safety scores are obtained without the motion tracker, which is not surprising as it is the slowest variant.

Question 4 tells us how well participants believe the following system performed. Our full system has the highest score of all, showing that every component of our system is crucial for the following accuracy. In particular, we notice how crucial our re-id module is as the lowest score is obtained by Ours\_w/o\_reid. We also observe the importance of using path planning and the benefits of combining both face and torso re-id models.

Finally, the users rated the smoothness of our system navigation in question 5. The low score obtained by Ours\_w/o\_pathplanning suggests that using path planning helps the system to navigate more smoothly.





%% file: tbl_tex/baselines.tex
\begin{table*}[t]
\centering
\begin{tabular}{c|ccccc}
\toprule
\multirow{2}{*}{Variants}       &        \multicolumn{5}{c}{\it components in our system}    \\
 & Motion Model & Torso-id Model & Face-id Model & Action by Visual Servo &  Action by Path Planning \\ \midrule
 Ours\_w/o\_reid~\cite{yoshimi2006development}   &    \cmark      & \xmark                    &    \xmark        &      \cmark               &    \cmark   \\
Ours\_w/o\_motion   &    \xmark      &    \cmark                       &     \cmark      &     \cmark               &   \cmark   \\ 
  Ours\_w/o\_torso      &     \cmark     &     \xmark                   &  \cmark          &       \cmark              &   \cmark     \\
  Ours\_w/o\_face~\cite{chen2017integrating}      &     \cmark     &     \cmark                   &  \xmark          &       \cmark              &   \cmark     
  \\
   Ours\_w/o\_visualservo      &     \cmark     &     \cmark                   &  \cmark          &       \xmark              &   \cmark     
  \\
   Ours\_w/o\_pathplanning      &     \cmark     &     \cmark                   &  \cmark          &       \cmark              &   \xmark    
   \\
  \midrule 
  Ours     &   \cmark     &     \cmark                   &  \cmark          &       \cmark              &   \cmark    \\
       \bottomrule
\end{tabular}
\caption{
\begin{flushleft}
Baseline methods studied in the paper for human following. We dissect our system and get rid of certain components to construct the variants.
\end{flushleft}}
\label{tbl:baselines}
\end{table*}

%% file: tbl_tex/results.tex
\begin{table*}[t]
\centering
\begin{tabular}{c|ccccc}
\toprule
\multirow{2}{*}{Variants}       &        \multirow{2}{*}{Avg. Speed (m/s)} & Avg. Following  & Avg. Distance & \multirow{2}{*}{\# of Losing Target} & \# of Following  \\
& & Distance (m) & to Obstacles (m) & & Wrong Person
\\
 \midrule \midrule 
 \multicolumn{6}{c}{\it Participants walking to random markers ($75$ trials) } \\
 \midrule \midrule 
 Ours\_w/o\_reid~\cite{yoshimi2006development}   &   0.87    & 1.3                   &   0.86       &     12 / 75             &   7 / 75   \\ 
Ours\_w/o\_motion   &    0.63    &  1.4                      &  0.81      &    5 / 75           &  2 / 75   \\ 
  Ours\_w/o\_torso      &  0.79    &   1.9                  & 0.79          &     9 / 75            & 2 / 75    \\
  Ours\_w/o\_face~\cite{chen2017integrating}      &    0.93     &     1.3             &  0.77        &       5 / 75           &  5 / 75     
  \\
   Ours\_w/o\_visualservo      &   0.71    &    1.5               &  1.15     &       4 / 75             &   1 / 75    
  \\
   Ours\_w/o\_pathplanning      &     0.97    &  1.7                   &  0.48       &       6 / 75       &  1 / 75   
   \\
  \midrule 
  Ours     &   0.92     &    1.2                   &  0.83          &       3 / 75              &   1 / 75    \\
 \midrule \midrule 
 \multicolumn{6}{c}{\it Participants following courses ($15$ trials) } \\
 \midrule \midrule 
 Ours\_w/o\_reid~\cite{yoshimi2006development}   &   1.16      & 1.6                    &    0.82       &    7 / 15              &   5 / 15  \\ 
Ours\_w/o\_motion   &    0.83     &  2.3                      &   0.78      &    4 / 15              &   1 / 15  \\ 
  Ours\_w/o\_torso      &    0.92     &     2.1                &  0.69         &       5 / 15            &   3 / 15    \\
  Ours\_w/o\_face~\cite{chen2017integrating}      &    1.19    &     1.4                 &  0.77         &       3 / 15             &  2 / 15    
  \\
   Ours\_w/o\_visualservo      &   1.03     &     1.7               &  0.91       &       3 / 15          &   0 / 15
  \\
   Ours\_w/o\_pathplanning      &  0.81     &     1.9                &  0.39     &       4 / 15             &  1 / 15   
   \\
  \midrule 
  Ours     &  1.24    &     1.5                  &  0.74        &       2 / 15             &   0 / 15   \\
       \bottomrule
\end{tabular}
\caption{
\begin{flushleft}
Quantitative results on 1) the average speed of agents when following human, 2) the average following distance of the agent, 3) the average distance to obstacles, 4) the number of losing the target person, and 5) the number of following a wrong person. We consider two conditions: 1) $5$ participants walking to $15$ random markers and 2) $5$ participants following $3$ different courses. For each trial, its environment contains obstacles, such as chairs, desks, and walls, and an interferer (trying to interfere the agent following the target person).
\end{flushleft}}
\label{tbl:results}
\end{table*}

%% file: tbl_tex/quetionairs.tex
\begin{table}[t]
\centering
\caption{
\begin{flushleft}
Quantitative results on questions for participants (see Section~\ref{subsec:metric_questions}). Each question is rated from $1$ to $10$, and $\uparrow$ means the higher the better; $\downarrow$ means the lower the better.
\end{flushleft}}
\begin{tabular}{c|ccccc}
\toprule
Variants        &        Q1($\uparrow$) & Q2($\downarrow$)  & Q3($\downarrow$) & Q4($\uparrow$) & Q5($\uparrow$)  
\\
 \midrule \midrule 
 \multicolumn{6}{c}{\it Participants walking to random markers ($75$ trials) } \\
 \midrule \midrule 
 Ours\_w/o\_reid~\cite{yoshimi2006development}   &    8.0      & 2.0                    &    2.0        &      4.0               &    7.0   \\ 
Ours\_w/o\_motion   &    9.3     &    2.0                       &     2.1      &     7.2               &   7.0   \\ 
  Ours\_w/o\_torso      &     8.0     &     2.0                   &  2.0          &      5.2              &   7.4     \\
  Ours\_w/o\_face~\cite{chen2017integrating}      &    8.0     &    2.0                   &  3.4          &    6.2             &  7.1     
  \\
   Ours\_w/o\_visualservo      &   7.0    &    2.1                   & 2.0          &    6.0             &   7.0    
  \\
   Ours\_w/o\_pathplanning      &     5.1    &    4.0                  &  6.0         &    5.3             &   4.6   
   \\
  \midrule 
  Ours     &  8.3    &   2.1                  &  3.4          &      8.2            &   7.5    \\
 \midrule \midrule 
 \multicolumn{6}{c}{\it Participants following courses ($15$ trials) } \\
 \midrule \midrule 
 Ours\_w/o\_reid~\cite{yoshimi2006development}   &   9.0     & 2.0                    &    2.2        &      3.4               &   8.0   \\ 
Ours\_w/o\_motion   &   9.0      &  2.0                       &   2.0     &   6.0            &   7.0   \\ 
  Ours\_w/o\_torso      &     9.0     &    2.3                   &  2.0          &      4.0          &   6.0     \\
  Ours\_w/o\_face~\cite{chen2017integrating}      &    9.2    &    2.4                  &  2.0          &    7.0             &   7.0    
  \\
   Ours\_w/o\_visualservo      &   9.0    &   2.0                  &  3.0          &   8.4            &  9.0     
  \\
   Ours\_w/o\_pathplanning      &     6.0   &     5.0                   &  5.2          &   4.4              &   5.1    
   \\
  \midrule 
  Ours     &   9.1    &   2.0                   & 2.0         &    7.0              &   8.3    \\
       \bottomrule
\end{tabular}
\label{tbl:quetionaires}
\end{table}

%% file: sections/conclusion.tex
\section{Conclusion}
\label{sec:conclu}

This paper addresses the issue of human following for mobile agents. We have observed that person re-identification is crucial for human following, particularly when the targeted individual is in a crowd or interacting with other people. To tackle this problem, we have developed a new person re-identification module that consists of three core components: 360-registration, identification models that use both face and torso, and motion tracker. We conducted a series of experiments and found that our approach outperforms previous methods, highlighting the effectiveness of each component in our person re-identification module. We believe that our work can help to improve the development of human following mobile agents and contribute to the advancement of artificial intelligence.